\documentclass[twoside,leqno,twocolumn]{article}

\usepackage[letterpaper]{geometry}

\usepackage{ltexpprt}
\usepackage{hyperref}
\usepackage{amsmath}
\usepackage{graphicx}

\begin{document}

\title{\Large CRD: Collaborative Representation Distance for Practical Anomaly Detection}
\author{Chao Han\thanks{hanc@mail.nwpu.edu.cn, Northwestern Polytechnical University}
\and Yudong Yan\thanks{yyanyanx12@doct.ub.edu, University of Barcelona}}

\date{}

\maketitle


\fancyfoot[R]{\scriptsize{Copyright \textcopyright\ 2024 by xxxx\\
Unauthorized reproduction of this article is prohibited}}





\begin{abstract} \small\baselineskip=9pt Visual defect detection plays an important role in intelligent industry.  Patch based methods consider visual images as a collection of image patches according to positions, which have stronger discriminative ability for small defects in products, e.g. scratches on pills.   However, the nearest neighbor search for the query image and the stored patches will occupy $O(n)$ complexity in terms of time and space requirements, posing strict challenges for deployment in edge environments.  In this paper, we propose an alternative approach to the distance calculation of image patches via collaborative representation models.  Starting from the nearest neighbor distance with $L_0$ constraint, we relax the constraint to $L_2$ constraint and solve the distance quickly in close-formed without actually accessing the original stored collection of image patches.  Furthermore, we point out that the main computational burden of this close-formed solution can be pre-computed by high-performance server before deployment. Consequently, the distance calculation on edge devices only requires a simple matrix multiplication, which is extremely lightweight and GPU-friendly.  Performance on real industrial scenarios demonstrates that compared to the existing state-of-the-art methods, this distance achieves several hundred times improvement in computational efficiency with slight performance drop, while greatly reducing memory overhead. 
\end{abstract}

\section{Introduction}
With the rise of Industry 4.0 and the Internet of Things (IoT), anomaly detection is becoming increasingly important in manufacturing\cite{2014A, 2022A}. Detecting and analyzing anomalies in real-time can help manufacturers ensure that only high-quality products are delivered to customers, reducing costs associated with product recalls and negative brand perception.

In this paper, we focus on visual failure detection, which utilizes computer vision algorithms to identify defects on object surfaces. Deep learning has revolutionized the field of computer vision,  with its ability to automatically extract features from large amounts of data, accurately identifying defects becomes possible. Existing works can be roughly divided into two categories: (1)Generative methods try to model the distribution of normal instances, i.e. Autoencoder-like methods reconstructing samples with its compressed representation\cite{2023Regularizing}, the intrinsic philosophy is that since the model only observed normal samples in training, it should return a small reconstruction error when facing to normal samples while a large one with abnormal samples. However, this hypothesis does not always hold true. Related researches declared that neural networks usually have too strong fitting ability to distinguish normals and outliers, that is, the decoder can reconstruct outliers well even though they were not observed during training\cite{2021Reconstruction}. (2)Another kind method is embedding based discriminative methods, which draw distinctions between normals and outliers based on a pre-defined feature space\cite{2016An}. Since Imagenet pretrained features have shown great potential on various downstream tasks\cite{imagenet_cvpr09}, it also becomes a default option for abnormal detection. By fixing the backbone, one can easily get down-sampled features and then analyse their statistics. The final distinguish principles are defined with patch-based score, e.g. the nearest-neighbor distance between a query patch and training patches, and the whole image score can be designed based on its sub-patch scores, e.g. the maximum. Embedding based discriminative methods consider each patch as a single unit and combinate them into together, it can alleviate the negative effects that the  reconstruction error of a whole image usually fails to detect small defects, this becomes especially important for production lines with mature technology, since there are only minor defects for common\cite{2022Application}.

Embedding-based methods depend on direct comparison between query patches and pre-saved (training) patches, therefore the size of pre-saved patch set has great influences for end-to-end abnormal detection, it will cause $O(n)$ demands of space and execution time, which brings challenges to the deployment on edge devices. To overcome this issue, several ways have been proposed. An intuitive idea is to model the statistical distribution of patches, and abnormals can be detected by artificially divided confidence interval\cite{Weixin2014Anomaly}. However, such methods exhibit performance degradation to some extent. Another way is downsampling patches, e.g. selecting a 'coreset' of patches instead of the full\cite{patchcore}, such methods are originated from the observation that image patches exhibit great similarity on production line scenarios. Not surprisingly, the selection of proportion greatly affects the final performance, and even after downsampling, the time and space costs are still unacceptable at some point.

Embedding-based solution implicitly partitions images into patches in the feature space, making failures that are inconspicuous on the entire image more prominent in the image patch. Beyond that, another key element is the nearest neighbor distance. Based on above analysis, we can summarize the following conclusions:
\begin{itemize}
	\item Learning-based distance metrics, whether complex neural network training with normal embeddings or simple GMM (Gaussian mixture model) estimation based on spatial positions, face the problem of overly strong learning ability, which is able to generalize to even the abnormal samples, leading to difficulty in distinguishing abnormal samples from normal ones, thus affecting the final detection accuracy.  
	\item Simple k-nearest neighbor distances can achieve optimal detection accuracy, but the process of calculating k-nearest neighbors is hardware-unfriendly, and saving too many training image patches requires a large memory overhead.
\end{itemize}
Based on the above analysis, we hypothesize that the key to visual anomaly detection is direct comparison between samples, rather than learning a complex model for distribution estimating. In this paper, we reformulate the nearest neighbor distance as a collaborative representation problem and relax the sparsity constraint to $L_2$, then propose a novel distance, named CRD (Collaborative Representation Distance), to improve the inference efficiency of anomaly detection models. Moreover, we further derive a close-formed solution based on CRD, achieve hundreds of times faster than nearest neighbor distance.

\section{Related Works}
\subsection{Embedding based anomaly detection}
Visual anomaly detection aims to identify and detect visual anomalies by analyzing abnormal patterns or unusual features in image or video data. Due to the unpredictability of anomalies, only normal samples exist during training, making it also known as one-class classification. An intuitive approach is to calculate the difference between the query sample and training set (consist of normal samples only), and further analyze whether the difference falls within the variance among normal patterns. Embedding based methods map samples to a feature space rather than the initial RGB space for difference calculation. Studies have shown that such solution can greatly reduce the noise introduced by imaging, and processing based on image patches makes it easier to identify minor defects, e.g. scratches on carpets often account for less than one percent of the total area.

SPADE (Semantic Pyramid Anomaly Detection\cite{spade}) is one of the pioneer works to apply embedding methods to anomaly detection. Using WideResnet\cite{wideresnet} model pretrained on Imagenet, it first maps normal images to feature space, in which the original images are compressed to smaller sizes and higher-dimensional channels based on feature pyramids.  For example, a picture with dimensions $H\cdot W \cdot3$ is mapped to a dimension of $\frac{H}{8} \cdot \frac{W}{8} \cdot 1536$, where $H$ and $W$ represent the height and width of the original image, respectively, and 3 \& 1536 represents feature channels. After mapping, the embedded representation has only $\frac{1}{64}$ $area$ since its height and width are $\frac{1}{8}$ of that before, and it can be considered that a vector on the feature space ($1\cdot1\cdot1536$) is responsible to an $8\cdot 8$ patch in the original RGB space since convolution operations have local receptive fields. Then save patch embeddings among all normal images to form a knowledge base. When a query image comes, transform it into the same feature space. For each vector corresponding patches, retrieve its k-nearest-neighbors from the knowledge base, compute the average distance between them as patch anomaly score. Finally, the image score can be determined by the maximum patch anomaly score since the whole product can be considered abnormal as long as part of it is abnormal. Despite its remarkable performance in detection accuracy, as the training set (normal images) grows larger, the knowledge base and execution time also increase. PaDiM (patch distribution model\cite{padim}) proposes to estimate the distribution of patch features according to their locations, so only the distribution parameters of each group need to be saved instead of an increasingly growing knowledge base. However, it is difficult to estimate the distribution of high-dimensional features, which damages its performance inevitably. PatchCore\cite{patchcore} reduces the size of the knowledge base by downsampling. A vital finding is that the representations in the knowledge base are highly redundant, which not only increases the computation and space demands but also damages the model accuracy. Therefore, PatchCore adopts coreset sampling to reduce redundancy while maintaining diversity in the knowledge base, making it state of the art currently.

Although PatchCore already achieves significant progress on the trade-off among accuracy-computation-memory, it is still challenging to deploy on edge devices. Regarding the selection of coresets, a fixed proportion of features are currently drawn from the knowledge base, which still does not break away from the $O(n)$ complexity limit. In this paper, we point out that the difference measure, nearest neighbor distance, is the key to obtains high accuracy but also brings large computations and memory costs. Consequently, we propose a new metric, named collaborative representation distance, to address this core issue.
\begin{figure*}[htb!]
	\centering
	\includegraphics[width=0.75\textwidth]{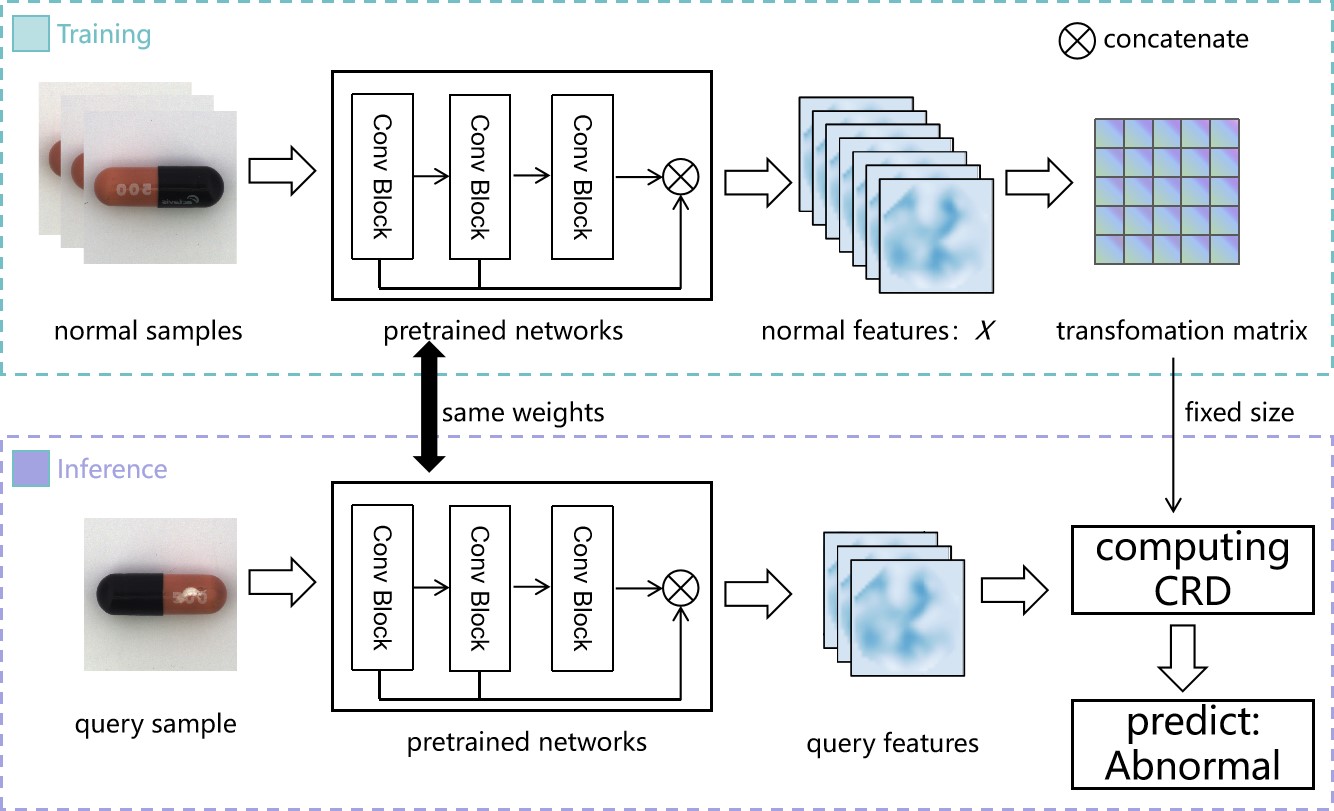}
	\caption{Graphical illustration of the overall framework.}
	\label{fig:framework}
\end{figure*}
\subsection{Collaborative Representation based Classifier}
CRC (Collaborative Representation based Classifier\cite{zhang2011sparse}) is first proposed for face recognition. Inspired by the wide-known SRC (Sparse Representation based Classifier\cite{4483511}), the authors explore the essential causes why SRC success in classification tasks, and claim that it is the collaborative representation, rather than $L_1$ constraints, helps SRC	to make correct predictions. Based on above analyses, CRC proposes to use $L_2$ constraints for solving collaborative representations quickly with close-formed solution. 

With the provided training samples($X = [x_1, x_2 \cdots x_n]$) and a query sample($y$), CRC first depicts the query samples via weighted sum of training set by minimizing the least square loss. Denoting the coefficients as $\rho$, the optimization can be write as:
\begin{equation}
\label{eq2.1}
min_{\rho_i\in \rho} L =  \  {||y-\sum_{x_i \in X}^{} x_i\cdot \rho_i||_2^2 + \lambda||\rho||_2^2 }
\end{equation}
where the first loss $||y-\sum_{x_i \in X}^{} x_i\cdot \rho_i||_2^2$ is reconstruction loss, which aims to make the reconstructed samples over the training set as close as possible to the query sample. The second loss is sparsity loss, which encourages the reconstructed coefficients $\rho$ to be sparse and $\lambda$ is used to balance these two losses. Since formula \ref{eq2.1} only concludes $L_2$ constraints, it can be solved by setting the partial derivative of $\rho$ to 0, that is:
\begin{equation}
\label{eq2.2}
\frac{\alpha L}{\alpha \rho} = \frac{1}{2}[(X\rho-y)+\lambda\rho] =  0
\end{equation}
\begin{equation}
\label{eq2.3}
\rho = (X^TX+\lambda I)^{-1}X^Ty
\end{equation}
Notice that vectorized expression of $X$ and $\rho$ is adopted in Formula \ref{eq2.2} - \ref{eq2.3}, which makes the result more concise. After obtaining the coefficients $\rho$, CRC fixes it and reconstructs the query sample via samples within specific categories. Intuitively, the category which has minimum reconstruction loss becomes the final prediction. 

\section{Methodology}
\subsection{Framework}
The overall framework is shown in Figure \ref{fig:framework}, which can be split into three key components, i.e. multi-scale feature extraction, CRD (Collaborative Representation Distance) computing and Inference. Feature extraction maps images from original color channels to intrinsic embedding space, with smaller sizes but higher dimensions. CRD computing provides a close-formed approximate solution for nearest neighbor distance, such that only a size-fixed matrix (whose size is determined by feature dimension) needs to be saved for inference, rather than an ever-growing knowledge base with normal samples. Inference gives the final prediction for query sample via CRD. For better understanding, we first formalize the problem and mathematical symbols: A typical visual anomaly detection problem is given a set of normal samples $X \in R^{d\cdot n} = [x_1, x_2 \cdots x_n]$ as training data, and we need to determine whether the query sample $y \in R^{d\cdot 1}$ belongs to the normal range, or it is considered as abnormal. And $d$ denotes the feature dimension of samples, for original images with RGB color channels, $d=3$. $n$ denotes the number of normal samples.

\subsection{Multi-Scale Feature Extraction}
Multi-scale feature extraction has been widely used in computer vision tasks, such as object detection and semantic segmentation\cite{8099589}. The philosophy behind it is that the receptive fields of features at different depths in convolutional neural networks are different, so multi-scale feature fusion can ensure good adaptability for objects of different sizes. For visual anomaly detection tasks, we do not make any assumptions about the size of defects, so multi-scale feature extraction is also necessary, which has been demonstrated by prior works. Drawing inspiration from FPN, a typical multi-scale feature extraction network upsamples deep features to the same dimension as shallow features and then fuses them by choosing either ADD or CONCATENATION operations. Therefore, we can actually construct multi-scale features flexibly according to latency and precision demands. To highlight the performance of CRD, we simply select the output features of Block2 and Block3 in WideResnet50 and concatenate them, like other methods do. Formally, we denote the feature set extract from $X$ as:
\begin{equation}
\label{eq3.1}
F = WideResnet50(X)
\end{equation}
It should be noted that as we later analyze features based on image patches, $F$ has larger dimensions than $X$, both in sample numbers and feature dimensions. In this paper, $F \in R^{d \cdot np}$, where $d$ is the feature dimension of multiscale WideResnet50 (which is equal to 1536 in this paper) and $p$ denotes the patch amount within an image.

\subsection{Collaborative Representation Distance}
Based on the analysis above, for visual anomaly recognition, geometric distances based nearest neighbor search in feature space can avoid overfitting issues of learning-based methods, but the calculation of geometric distances requires storing feature sets, which is not efficient in terms of time and space. So the goal is: First, no need to save features. Second, do not have too much fitting ability.
\subsubsection{Reformulate Nearest Neighbor Distance}
Nearest neighbor distance first calculates the geometric distance between the query sample and feature set, then selects the minimum distance as the final metric. In fact, it can be considered as a minimization problem under $L_0$ constraint.

\begin{align}
\label{eq3.2}
D_{nn} &= min\enspace ||y-\sum_{\rho=1}^{card(F)}\rho_iF_i||_2^2\\
\nonumber &s.t.\enspace ||\rho||_0 = 1
\end{align}
By constraining the $L_0$ norm of $\rho$ to be 1, we can ensure that only the smallest geometric distance is selected, which is equivalent to the nearest neighbor actually. Similarly, we can constrain the norm of $\rho$ to be $k$ to obtain $k$-nearest neighbor distances. 

\subsubsection{Relax the Constraint to $L_2$}
By reformulating the calculation of nearest neighbor distance as reconstruction based on feature set, L0 constraint becomes a new obstacle since it is hard to optimize. Inspired by the success of SRC and CRC, we can conclude that $L_1$ and $L_2$ relaxation are both effective substitutions for $L_0$ constraint. So naturally we further assume that for visual anomaly detection, $L_2$ relaxation does not affect the accuracy too much. In conjunction with the Lagrange multiplier method, equations \ref{eq3.2} can be rewritten as:
\begin{align}
\label{eq3.3}
D_{cr} &= min_\rho\enspace ||y-\rho F||_2^2 + \lambda||\rho||_2^2
\end{align}
Now we can see that the distance has similar forms to the process of solving coefficients in CRC. The difference between them lies in the subsequent processing of the coefficients $\rho$. As a standard classification problem, CRC needs to calculate the reconstruction error separately for different classes of objects, while we need it under all samples since they all belong to the same class, i.e. normal.

\subsubsection{Solution of CRD}
Equation \ref{eq3.2} states a minimization problem under quadratic constraints, we first solve $\rho$ by setting its partial derivative to zero.
\begin{align}
\label{eq3.3}
\nonumber \frac{\partial D}{\partial \rho} &= \frac{1}{2}[F^T(F\rho-y)]+\frac{1}{2}\lambda\rho = 0\\
&\Rightarrow \rho = (F^TF+\lambda I)^{-1}F^Ty
\end{align}
where $I_{np}$ denotes the $np$-th order identity matrix and $\{\}^{-1}$ denotes the inverse matrix. Furthermore, substitute equation \ref{eq3.3} to the reconstruction term of equation \ref{eq3.2}.
\begin{align}
\label{eq3.4}
\nonumber D_{cr} &= ||F((F^TF+\lambda I)^{-1}F^Ty)-y||_2^2\\
 &= [F(F^TF+\lambda I)^{-1}F^T-I]\cdot y
\end{align}
It is worth noting that the left side of the equation (denoting with $M_{pre} = F(F^TF+\lambda I)^{-1}F^T-I$) is completely independent of y, so we can compute it on high performance servers before deploying. Further based on the calculation of vector dimension, we have $M_{pre}\in R^{d\cdot d}$, which illustrates that the size of $M_{pre}$ is completely independent with that of feature set, makes it totally different with current feature set downsampling strategies.

\subsection{Inference}
For the query sample $x$, we first map it to a set of feature vectors corresponding to different positions in the image via the same feature extraction network ($WideResnet50$). Then, the anomaly score is computed by CRD using equation \ref{eq3.4}. It is worth noting that we can calculate the CRD for the full set of vectors at once using General Matrix Multiplication (GEMM), which is executed efficiently for modern accelerators such as General-Purpose Graphics Processing Unit (GPGPU) or Tensor Processing Unit (TPU). In contrast, traditional nearest neighbor distance requires calculating the maximum value for vectors at different positions separately, which is inefficient since it cannot be highly parallelized. The final anomaly score is determined by a pre-defined threshold, which can be learned by normal samples, i.e. the biggest CRD within normal samples.

\section{Experiments}
\begin{table*}[htb!]
	\centering
	\caption{Image-Level AUC}
	\begin{tabular}{|c||cccccc||c|}
		
		\hline
		Method& STFPM &PaDiM & CFA    & CFlow    & EfficientAd & PatchCore &CRD  \\
		\hline
		\hline
		Carpet & 95.7 & \textbf{99.5} & 97.8 & 98.6 & 97.2 & 98.4 & 98.9\\
		Grid & 97.7 & 94.2 & 96.1 & 96.2 & \textbf{99.8} & 95.9 & 98.2\\
		Leather & 98.1 & \textbf{100.0} & 99.0 & \textbf{100.0} & \textbf{100.0} & \textbf{100.0} & \textbf{100.0}\\
		Tile & 97.6 & 97.4 & 99.9 & 99.9 & 99.9 & \textbf{100.0} & \textbf{100.0}\\
		Wood & 93.9 & 99.3 & \textbf{99.4} & 99.3 & 98.4 & 98.9 & 98.5\\
		Bottle & 98.7 & 99.9 & 99.8 & \textbf{100.0} & 99.1 & \textbf{100.0} & \textbf{100.0}\\
		Cable & 87.8 & 87.8 & 97.9 & 89.3 & 94.5 & \textbf{99.0} & 93.8\\
		Capsule & 73.2 & 92.7 & 87.2 & 94.5 & 95.7 & \textbf{98.2} & 97.5\\
		Hazelnut & 99.5 & 96.4 & \textbf{100.0} & \textbf{100.0} & 94.8 & \textbf{100.0} & 96.8\\
		Metal Nut & 97.3 & 98.9 & \textbf{99.5} & \textbf{99.5} & 98.9 & 99.4 & 98.0\\
		Pill & 65.2 & 93.9 & \textbf{94.6} & 92.4 & 92.6 & 92.4 & 91.0\\
		Screw & 82.5 & 84.5 & 70.3 & 90.8 & \textbf{97.5} & 96.0 & 86.8\\
		Toothbrush & 50.0 & 94.2 & \textbf{100.0} & 89.7 & \textbf{100.0} & 93.3 & 90.8\\
		Transistor & 87.5 & 97.6 & 95.7 & 94.3 & 96.5 & \textbf{100.0} & \textbf{100.0}\\
		Zipper & 89.9 & 88.2 & 96.7 & 98.4 & 97.1 & 98.2 & \textbf{99.3}\\
		\hline
		AVG&87.6 & 95.0 & 95.6 & 96.2 & 97.5 & \textbf{98.0} & 96.6\\
		\hline
		
	\end{tabular}
	\label{res1}
\end{table*}
\begin{table*}[htb!]
	\centering
	\caption{Pixel-Level AUC}
	\begin{tabular}{|c||cccccc||c|}
		
		\hline
		Method& STFPM &PaDiM & CFA    & CFlow    & EfficientAd & PatchCore &CRD  \\
		\hline
		\hline
		Carpet & 98.7 & \textbf{99.1} & 98.0 & 98.6 & 94.8 & 98.8 & 98.7\\
		Grid & \textbf{98.9} & 97.0 & 95.4 & 96.8 & 93.7 & 96.8 & 97.4\\
		Leather & 98.0 & \textbf{99.3} & 98.9 & \textbf{99.3} & 97.6 & 99.1 & 99.0\\
		Tile & 96.6 & 95.5 & \textbf{98.5} & 96.8 & 90.6 & 96.1 & 95.8\\
		Wood & 95.6 & 95.7 & \textbf{97.4} & 92.4 & 86.7 & 93.4 & 93.3\\
		Bottle & 96.6 & 98.5 & \textbf{98.9} & 98.1 & 97.6 & 98.4 & 98.0\\
		Cable & 91.3 & 97.0 & \textbf{98.8} & 95.5 & 98.6 & \textbf{98.8} & 96.8\\
		Capsule & 95.6 & 98.8 & \textbf{98.9} & 98.8 & 95.7 & 98.8 & 98.8\\
		Hazelnut & 97.4 & 98.5 & 98.5 & \textbf{99.0} & 97.7 & 98.7 & 97.8\\
		Metal Nut & 96.1 & 98.2 & \textbf{99.2} & 98.2 & 98.4 & 98.9 & 98.6\\
		Pill & 94.6 & 96.6 & \textbf{98.8} & 98.3 & 97.8 & 98.0 & 98.1\\
		Screw & 98.8 & 98.8 & 97.9 & 97.9 & 98.6 & \textbf{98.9} & 95.8\\
		Toothbrush & 17.8 & \textbf{99.1} & \textbf{99.1} & 98.5 & 96.4 & 98.8 & 98.7\\
		Transistor & 80.7 & 97.6 & 97.7 & 89.7 & 94.7 & \textbf{98.1} & 96.5\\
		Zipper & 98.0 & 98.6 & \textbf{99.0} & 98.0 & 96.0 & 98.3 & 98.3\\
		
		\hline
		AVG& 90.3 & 97.9 & \textbf{98.3} & 97.1 & 95.7 & 98.0 & 97.4\\
		\hline
		
	\end{tabular}
	\label{res2}
\end{table*}
\begin{figure}[htb!]
	\centering
	\includegraphics[width=0.45\textwidth]{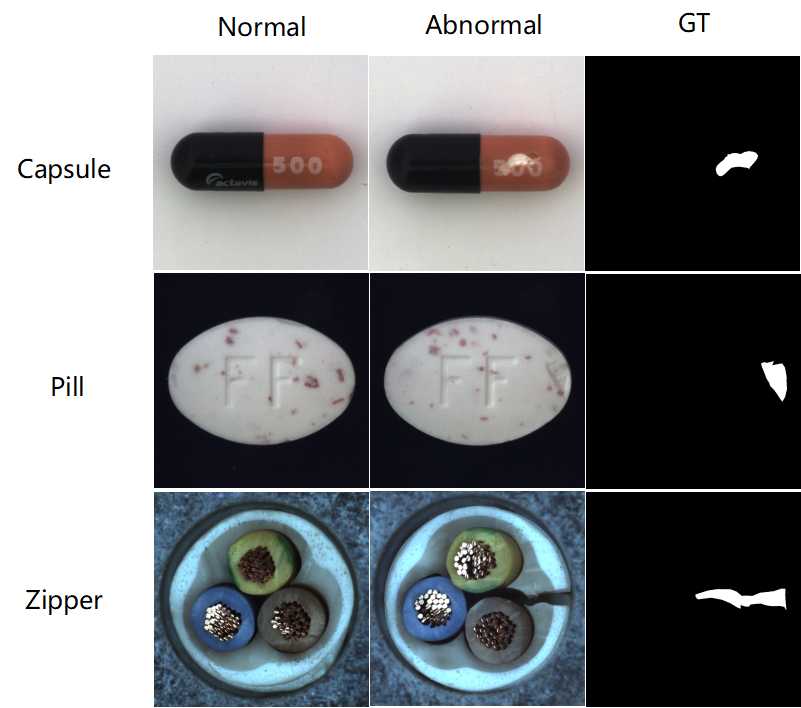}
	\caption{Examples visulization in MvTec Ad.}
	\label{fig:mvtec}
\end{figure}
\subsection{Datasets}
\textbf{Mvtec AD} dataset provides up to 15 kinds of visual defects in real industrial production, including \textit{Carpet, Grid, Leather, Tile, Wood, Bottle, Cable, Capsule, Hazelnut, Metal Nut, Pill, Screw, Toothbrush, Transistor, Zipper}. There are 3629 images used for training (containing only normal samples) and 1725 for validation. In addition to the whole image labels (normal or abnormal), it also provides pixel-level annotations, while for embedding-based methods we can also obtain pixel-level predictions by considering whether the feature is abnormal at each location. Therefore, we will compare both image-level and pixel-level metrics in our experiments. Figure \ref{fig:mvtec} visualizes three categories of samples.
\subsection{Settings}
In this paper, we compare the proposed method (denotes as \textbf{CRD}) with several state-of-the-art works under the same settings. 
\begin{itemize}
\item{\textbf{STFPM}\cite{stfpm}}: trains a student model with only normal data, computes the difference of student and teacher model as anomaly scores.
\item{\textbf{PaDiM}\cite{padim}}: models each position with Multivariate Gaussian Distribution. 
\item{\textbf{CFA}\cite{cfa}}: adapts target features with coupled-hypersphere.
\item{\textbf{CFlow}\cite{cflow}}: introduces conditional normalizing flows for anomaly detection.
\item{\textbf{EfficientAd}\cite{efficient}}: detects anomalies efficiently via a global autoencoder.
\item{\textbf{PatchCore}\cite{patchcore}}: downsamples feature set for efficient inference.
\end{itemize}
For fairness, all methods adopt $WideResnet50$ for feature extraction except EfficientAd, since its network architecture designing,  Patch description network (PDN), is a key contribution. We follow the implementation of Anomalib, provided by $OpenVINO^{TM} Toolkit$\footnote{https://github.com/openvinotoolkit}, and apply the standard training \& evaluating pipeline for all the methods. About the parameter setting, we adopt the official reported performance for compared methods, and set $\lambda = 5$ for our CRD. AUC (Area Under the ROC Curve) is used to evaluate the performance.

\subsection{Results}
\begin{figure}[htb!]
	\centering
	\includegraphics[width=0.45\textwidth]{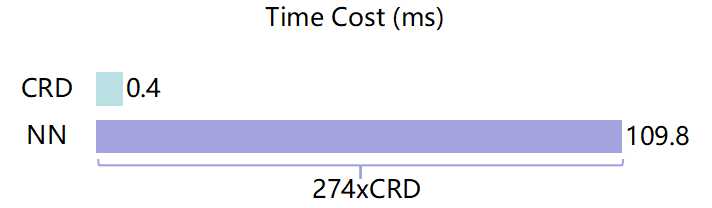}
	\caption{Time cost for CRD and NN.}
	\label{fig:time}
\end{figure}

The image-level and pixel-level AUC can be found in Table \ref{res1} and \ref{res2}. In summary, the proposed CRD has comparable performance with state-of-the-art works whether in anomaly detection or locazation. For anomaly detection, CRD achieves first place in 6 / 15 tasks and 3 second place. CRD obtains 96.6\% average AUC score in 15 tasks, compared to the best-performed PatchCore (98.0\%), only 1.4\% AUC drop. For anomaly detection, we can see that the gap is further narrowed, CRD achieves 97.4\% in average pixel-level AUC, with only a 0.9\% drop in AUC score compared to the best method (CFA, 98.3\%).

Rather than accuracy, the superiority of the proposed CRD is presented in the running speed and memory. We record the running time of distance computing between PatchCore and CRD on the same machine with single tesla V100 GPU, the results are shown in Figure \ref{fig:time}. Not surprisingly, CRD runs hundreds of times (274x in average for 15 tasks) faster than nearest neighbor search. It should be noted that PatchCore has downsampled the feature set, and only 1\% of the features are selected as the core set. Based on the linear complexity calculation, CRD is more than 20,000 times faster than the nearest neighbor distance on the full feature set, and this value increases with the size of the training set getting larger. As for memory saving, we have illustrated that the pre-computed matrix is a fix-shaped model about training set, its shape is determined by feature extraction models. When applied the downsampling in PatchCore, we find that the number of training set is on the same order of magnitude as feature dimensions, so there are only about 35\% memory savings on Mvtec Ad. But it still needs to be emphasized that CRD has more scaling ability for large data sets, since the memory it occupies is training set independent.

\subsection{Parameter Sensitivity About $\lambda$}
\begin{table}[htb!]
	\centering
	\caption{Image-Level AUC}
	\begin{tabular}{|c||ccccc|}
		
		\hline
		$\lambda$ & 0.1 & 1  &3   & 5 & 10  \\
		\hline
		AUC(\%) & 93.1 & 93.6 &95.1 & \textbf{96.6} & 94.7  \\
		\hline	
	\end{tabular}
	\label{res3}
\end{table}

In this paper, $\lambda$ is responsible for controlling the complexity of the reconstruction coefficient $\rho$. As it increases, $\rho$ will become more sparse, but at the same time, the reconstruction loss may increase correspondingly. In addition, to solve $\rho$, the matrix inversion of $F^TF$ needs to be calculated. When there are too few samples, the matrix may not be full-rank since that normal features are very likely to be linearly dependent, so introducing $\lambda$ can also help stabilize the solution. Table \ref{res3} shows the image-level AUC scores over a set of $\lambda$, as it increases, model's performance improves and then drops, which is consistent with our analysis.

\section{Conclusion}
In this paper, we propose a novel metric for visual anomaly detection. Inspired by the success of sparse representation in face recognition, we extend the widely-used nearest neighbor distance to collaborative representation distance, and derive the close-formed solution, i.e. only a fixed-size matrix need to be saved when deploying, rather than an increasing-size codebook. Furthermore, the full computing can be intergrated into a matrix multiplication, which is highly parallelized and optimized well with modern GPU or TPU devices. Experiments on 15 real scenarios show that the proposed method is hundreds of times faster than nearest neighbor distance with acceptable loss of performance.

\bibliographystyle{siam}
\bibliography{refref}
\end{document}